\documentclass[a4paper,11pt,twocolumn,twoside]{article}
\usepackage[dvips]{graphicx}
\usepackage{sepln_en}
\usepackage{fullname}
\usepackage[utf8]{inputenc}

\input epsf

\usepackage[utf8]{inputenc} 
\usepackage[T1]{fontenc}    
\usepackage{url}            
\usepackage{bm}            
\usepackage{mathtools}
\usepackage{booktabs}       
\usepackage{amsfonts}       
\usepackage{microtype}      
\usepackage{lipsum}
\usepackage{graphicx}
\graphicspath{ {./images/} }
\usepackage[LGR,T1]{fontenc}
\usepackage[utf8]{inputenc}   
\usepackage[dvipsnames]{xcolor}
\usepackage{changes}
\usepackage{lipsum}
\usepackage{listings}

\begin{document}

\title{Open Conversational LLMs do not know most Spanish words}

\author {\textbf{Javier Conde,$^1$} \textbf{Miguel Gonz\'alez,$^1$} \textbf{Nina Melero,$^{1,4}$} \textbf{Raquel Ferrando,$^1$}  \textbf{Gonzalo Mart\'inez,}$^2$  \\  
\textbf{Elena Merino-G\'omez,$^3$} \textbf{Jos\'{e} Alberto Hern\'{a}ndez,$^2$} \textbf{Pedro Reviriego$^1$} \\ 
$^1$ETSI de Telecomunicaci\'on, Universidad Polit\'{e}cnica de Madrid \\  
\textit{email: \{javier.conde.diaz,miguel.gonsaiz,i.melero,pedro.reviriego\}@upm.es,} \\ \textit{raquel.ferrando@alumnos.upm.es} \\   
$^2$Universidad Carlos III de Madrid \\  
 \textit{email: jahgutie@it.uc3m.es,gonzmart@pa.uc3m.es}\\
$^3$Universidad de Valladolid \\
\textit{email: elena.merino.gomez@uva.es}\\
$^4$New York University-Madrid \\
\textit{email: mim8070@nyu.edu} \\
}

\seplntranstitle{Los LLM conversacionales \textcolor{black}{abiertos} no reconocen la mayoría de las palabras del español}

\seplnclave{LLM, vocabulario, \textcolor{black}{modelos abiertos}, español.}

\seplnresumen{La popularidad de los grandes modelos de lenguaje, o LLM  del inglés \textit{Large Language Models}, con los que los usuarios pueden interactuar ha llevado al desarrollo de un gran número de modelos \textcolor{black}{abiertos}. Estos modelos se evalúan con múltiples conjuntos de pruebas para valorar sus capacidades para responder preguntas o resolver problemas sobre casi cualquier tema posible, o para probar su habilidad para razonar o interpretar textos. Sin embargo, la evaluación del conocimiento que estos modelos tienen de los idiomas ha recibido mucha menos atención. Por ejemplo, las palabras que pueden reconocer y usar en diferentes idiomas. En este artículo evaluamos el conocimiento que los LLM conversacionales \textcolor{black}{abiertos} tienen de las palabras en español utilizando una muestra de palabras de un diccionario de referencia. Los resultados muestran que los LLM conversacionales \textcolor{black}{abiertos} producen significados incorrectos para una fracción importante de las palabras y no son capaces de usar la mayoría de las palabras correctamente para escribir frases con contexto. Estos resultados muestran cómo el español se queda atrás en la carrera de los LLM de código abierto y destacan la necesidad de impulsar la equidad lingüística en los LLM conversacionales asegurando que proporcionen un rendimiento similar en todos los idiomas.}

\seplnkey{LLMs, Vocabulary, Open \textcolor{black}{models}, Spanish.}

\seplnabstract{The growing interest in Large Language Models (LLMs) and in particular in conversational models with which users can interact has led to the development of a large number of \textcolor{black}{open} chat LLMs. These models are evaluated on a wide range of benchmarks to assess their capabilities in answering questions or solving problems on almost any possible topic or to test their ability to reason or interpret texts. Instead, the evaluation of the knowledge that these models have of the languages has received much less attention. For example, the words that they can recognize and use in different languages. In this paper, we evaluate the knowledge that \textcolor{black}{open} chat LLMs have of Spanish words by testing a sample of words in a reference dictionary. The results show that \textcolor{black}{open} chat LLMs produce incorrect meanings for an important fraction of the words and are not able to use most of the words correctly to write sentences with context. These results show how Spanish is left behind in the  LLM race and highlight the need to push for linguistic fairness in conversational LLMs ensuring that they provide similar performance across languages.}



\maketitle

\clearpage

\linespread{1.0}


\section{Introduction}
\label{sec:introduction}

The development of Large Language Models (LLMs) with billions of parameters and trained with huge amounts of text has pushed the performance of computer-based Natural Language Processing (NLP) to new limits achieving impressive performance in a wide range of tasks such as summarization, translation, retrieval of information, and conversation \cite{LLMreview}. These LLMs are the foundation for systems that can interact with humans, such as the well-known ChatGPT \cite{ChatGPTOverview}, which is now used by hundreds of millions of people. These conversational LLM-based tools are being used to build autonomous agents that can perform many tasks interacting in an environment with humans and also with other agents \cite{autonomousagents}. These agents will soon replace traditional ChatBots \cite{ChatBots} enabling a myriad of applications, from a personal assistant that manages our daily office tasks to a teaching assistant or a customer support agent.  

As LLM-based tools and agents are adopted and eventually become widespread, the research community is focusing on evaluating their performance in a wide range of tasks and topics. Those include maths \cite{Mathmeasuring}, reasoning \cite{CommonSensemeasuring} but also comprehensive tests on a large number of tasks and topics \cite{MultipleTasksmeasuring},\cite{BIGMeasuring}. Instead, the linguistic aspects of the text generated by LLM-based tools have received much less attention, with only a few works exploring the linguistic features \cite{HCC3}, phonological bias \cite{HCC2} or lexical \textcolor{black}{diversity}  \cite{reviriego2023playing} of the AI-generated text. However, this is important as after all LLMs will likely account for a significant fraction of the text generated in the future. 

In addition to the linguistic features, it is also of interest to understand the knowledge that LLMs have about the different languages, as they have been optimized in most cases for English. This may lead to a situation in which agents and other AI tools have worse performance for other languages, leading to an unfair scenario. This unfairness is twofold, firstly it will reinforce the dominant languages and secondly, it will put non-native English speakers at a disadvantage. There have been different efforts to develop truly multilingual LLMs such as BLOOM \cite{workshop2022bloom} or translation tools \cite{NLLB} but the performance of AI tools and models is still highly dependent on the language, as shown in \cite{LNLB} for AI text to image generators.    

To understand the importance given to different languages, a simple approach is to look at the amount of text used for training in each language. For example, in GPT-3 more than 181 billion words in English were used for training compared to only 1.5 billion in Spanish, representing 92.65\% and 0.77\% of the training set, respectively. This shows the prevalence of English in the training phase and more so if we take into account that Spanish is the fourth language with more words in the training dataset\footnote{\url{https://github.com/openai/gpt-3/blob/master/dataset_statistics/languages_by_word_count.csv}}. The same trend is observed on other commercial models such as PaLM2 with only 2.11\% of Spanish compared to 77.98\% in English \cite{chowdhery2023palm}. For open models, the situation is similar or even worse, for example, in LLaMa-2 only 0.11\% of the training set is in Spanish \cite{touvron2023llama2}. In fact, for LLaMa models a recent work suggests that they work internally in English \cite{wendler2024llamas}. 

The dominance of a single language in the training dataset of LLMs suggests that performance will be worse for other languages although the evaluation results, for example, of GPT-4 show a very similar performance for English and Spanish on the Measuring Massive Multitask Language Understanding (MMLU) benchmark \cite{MultipleTasksmeasuring} with an accuracy of 85.5\% and 84\% respectively \cite{GPT4}. However, the performance of LLMs in languages other than English and in some cases Chinese has received little attention compared to the amount of evaluation performed for the dominant languages. Therefore, both linguistic features and languages have not been the focus of the LLM evaluation to date. 

One aspect that is of interest is how LLMs use the vocabulary of a language and the fraction of words that they can recognize \cite{ChatWords}. On one hand, the vocabulary of LLMs may influence how languages evolve in the future, as words not used by LLMs may eventually be less and less used. On the other hand, the fraction of words in a language that LLMs can recognize can be an indication of the degree of lexical knowledge that they have in different languages. 

In this work, we contribute to the study of the lexical knowledge of LLMs by evaluating a representative group of open conversational LLMs from several companies and organizations and with different sizes on a sample of the words present in a reference Spanish dictionary \cite{seco1999diccionario}. 

The rest of the paper is organized as follows, in section \ref{sec:motivation} we present the motivation and objectives of the work. Section \ref{sec:models} discusses the selection of the open conversational LLMs for evaluation and the evaluation methodology is described in section \ref{sec:methodology}. The results are presented in section \ref{sec:results} followed by a discussion of the main insights as well as ideas for continuing this work in section  \ref{sec:discussion}. The paper ends with the conclusion in section \ref{sec:conclusions}.

\section{Motivation and objectives} 
\label{sec:motivation}

As discussed in the introduction, there is a lack of studies that analyze the linguistic aspects and, in particular, the lexical knowledge of LLMs and also of their understanding of languages other than English. This motivates our work, that tries to address both issues by studying the lexical knowledge of Spanish in open conversational LLMs. The rationale behind focusing on conversational LLMs is twofold. Firstly, they lend themselves to testing by asking the models directly if they know the meaning of words and if they are capable of using them meaningfully. Secondly, because they are widely used in many tools and agents. The choice of the language is given by us being native Spanish speakers but also because Spanish is typically in the top five languages with most words in the training set. This means that the results obtained will probably be better than for most other languages that have less data in the training set. Finally, we focus on open LLMs as there is a large number of models, with different sizes, and architectures, and even some of them are adapted for Spanish which enables a better evaluation of the impact of the different parameters on lexical knowledge. The evaluation of open models is also intended to provide feedback to the open-source community that could be used to improve the lexical knowledge of new conversational LLMs across different languages. 


We set the following objectives in our study: 

\begin{enumerate}
    \item Evaluate the fraction of Spanish words that open conversational LLMs can recognize and use in context.
    \item Provide an overview of the current lexical knowledge of Spanish in open conversational LLMs.    
    \item Analyze the impact of model size on the knowledge of the words.
    \item Compare the knowledge of models designed to support multiple languages to those focused on English and Chinese.
    \item Analyze whether the adaptation of the model to Spanish from a pre-trained model improves the knowledge of the words.
\end{enumerate}

The methodology to test the conversational LLMs and evaluate the responses is described in the next section. The results can be processed to elaborate a summary of the lexical knowledge of conversational LLMs and how it depends on different factors. For example, the dependency on the model size can provide insights into whether acquiring the lexicon is limited by the training set, by model size, or possibly by both. Similarly, it is of interest to understand if models that are specifically designed to support several languages \cite{workshop2022bloom} have a better knowledge of the Spanish vocabulary than models that are optimized for one or two languages\footnote{For example \url{https://huggingface.co/01-ai/Yi-34B-Chat}.}. The same reasoning applies to the first conversational LLMs that are adapted to improve their performance in Spanish, do they have a better knowledge of the Spanish lexicon? Next, we discuss the selection of the conversational LLMs and the evaluation methodology used to pursue our objectives.

\section{Models evaluated} 
\label{sec:models}

There is a large (and growing) number of open models that have been designed to interact with users\footnote{See for example at \url{https://huggingface.co/spaces/lmsys/chatbot-arena-leaderboard}}. For example, filtering the models available at Huggingface to select only the ones for natural language processing in the category conversational returns more than 2,700 results. Although some of those may be the same model on different formats or with different quantization, the number of conversational LLMs is clearly too large to make an exhaustive evaluation feasible. Therefore, the first step is the selection of a subset of conversational LLMs for evaluation. 

The number of models in the evaluation set has to be manageable so that evaluating the models requires a reasonable effort but at the same time, it captures different types and sizes of conversational LLMs. In the selection, we focus on models that are widely used and with different sizes and target languages. 

The evaluation set includes three LlaMa-2 \cite{touvron2023llama2} models from Meta with different sizes ranging from 7B to 70B and two models from Mistral AI with 7B \cite{jiang2023mistral} and 46.7B \cite{jiang2024mixtral} parameters. LlaMa and Mistral models have been used as a starting point to build other models by fine-tuning or adaptation such as Zephyr-7B \cite{tunstall2023zephyr}, OpenChat \cite{wang2023openchat} or WizardLM \cite{xu2023wizardlm}. Therefore, the insights gained on LLaMa and Mistral models may be applicable to their derivative models. Solar10.7B \cite{kim2023solar} which improves Mistral-7B using a scaling technique to derive a larger model with close to 11B parameters is also included in our evaluation set to check if scaling has any impact on lexical knowledge. Two models recently introduced by 01.AI and optimized for English and Chinese\footnote{\url{https://github.com/01-ai/Yi}} with 6B and 34B that have shown good performance in several benchmarks are also included in the set. This first group of general models that are mostly targeted towards English and in some cases also Chinese is completed with Gemma-7B \cite{gemma_2024} which has been recently released by Google. 

 
A second group of models that are either multilingual by design or that have been optimized for Spanish is also included in the evaluation set. The goal is to check if these optimizations improve the lexical knowledge of Spanish. Three models are evaluated, the first two Bloomz-7B1\footnote{\url{https://huggingface.co/bigscience/bloomz-7b1}} (a model that has been trained specifically to support several languages) and Flor-6.3BInstructed\footnote{\url{https://huggingface.co/projecte-aina/FLOR-6.3B-Instructed}} (a model optimized for Spanish)  are derived from Bloom7B, a multilingual open model \cite{workshop2022bloom}.  The third, Bertin-6B is a version of GPT-J 6B fine-tuned for Spanish \cite{BERTIN-GPT}. These three models enable an evaluation of the changes in lexical knowledge introduced by optimizing the models for Spanish.

The twelve models evaluated and their main features are summarized in Table
\ref{tab:models}. The third column shows the quantization used to run the models in terms of bits per parameter. The parameters are available with 32-bit precision for most models and 32-bit operations are also supported by the GPUs. Therefore, when possible 32 bits are used, when there is not enough memory fewer bits are used. It can be observed that the LLMs selected cover a wide range of model sizes from 7 to 70 billion parameters from different companies. The quantization of the model parameters has been selected so that they can be run on a single GPU in our computing cluster\footnote{We use NVIDIA A100 GPUs with 40GB of memory and RTX-A6000 GPUs with 48GB of memory.}.


\begin{table*}[h]
    \centering
    \begin{tabular}{|c|c|c|}
    \hline
    Model & Parameters & Bits/Parameter    \\
    \hline
    Llama-2-7b-chat-hf & 6.7B & 32    \\  
    Llama-2-13b-chat-hf & 13B & 16   \\ 
    Llama-2-70b-chat-hf & 69B & 4   \\ 
    \hline
    Mistral-7b-Instruct & 7.2B & 32  \\ 
    Mixtral-8x7b-Instruct & 46.7B & 4  \\ 
    \hline
    Gemma-7b-it & 8.54B & 32    \\ 
    \hline    
    SOLAR-10.7b-Instruct & 10.7B & 16   \\ 
    \hline    
    Yi-6b-Chat & 6B & 32   \\ 
    Yi-34b-Chat & 34.4B & 8   \\ 
     \hline    
    Bloomz-7b1  & 7.1B & 32  \\ 
    \hline    
    FLOR-6.3b-Instructed  & 6.3B & 32  \\ 
    \hline    
    Bertin-6b & 6B & 32   \\ 
    \hline    
    \end{tabular}
    \caption{Conversational LLMs considered in the evaluation.}
    \label{tab:models}
\end{table*}

\section{Evaluation methodology} 
\label{sec:methodology}

This section describes the methodology used to evaluate the lexical knowledge of the conversational LLMs covering how to ask the models, what to ask in terms of words, and the processing and interpretation of the responses.

\subsection{Prompts}

Since all the models evaluated are conversational, we can ask them about the words directly using different prompts to assess if they know a given word and if they can use it meaningfully. The most straightforward procedure is to ask the models directly if they know the word or if it is correct. However, this may be misleading as LLMs are known to suffer hallucinations and provide inconsistent responses \cite{hallucinationsReview}. Therefore, we initially use two prompts that ask the LLMs for the meaning of the word and to use the word to write sentences. In more detail, the prompts (and their English translations) used are:

\begin{itemize}
    \item \textbf{Prompt A (meaning)}: ``\textit{Dime la definición de la palabra `<word>'.}'' (``Write the definition of the word  `<word>' '')
    \item \textbf{Prompt B (use)}: ``\textit{Escribe dos frases, una con la palabra `<word>', y otra que no contenga esa palabra, pero que esté relacionada con la primera y complemente su significado.}'' (``Write two sentences, one with the word `<word>', and another that does not contain that word, but that is related to the first and complements its meaning.'')
\end{itemize}

The first one is intended to assess if the model knows the meaning of the word and the second is to check if the model can use the word in a meaningful way. These prompts are based on assessment methods for human learners informed by Communicative Language Teaching, integrating the lexical approach \cite{TwoSentencesb}. The prompts are designed as ‘tasks’; in this case, small-scale written production exercises which ask for the generation of context. The ability to recognise and produce individual items will not suffice to complete these tasks successfully: in order to do so, the testee will rather have to create and use language as a functional part of a text. This integrative method of vocabulary assessment is aimed at evaluating comprehension, fluency, and accuracy, all key aspects of lexical competence \cite{TwoSentences}.


In addition to those two prompts with open answers, we test three additional prompts in which the models are asked to answer only ``Yes'' or ``No''. The selected prompts and their English translations are as follows: 

\begin{itemize}
    \item \textbf{Prompt 1}: ``\textit{¿Conoces el significado de la palabra `<word>'? Responde solo `Sí' o `No' y sé sincero, por favor.}'' (``Do you know the meaning of the word `<word>'? Answer only `Yes' or `No' and be honest, please.'')
    \item \textbf{Prompt 2}: ``\textit{¿Existe la palabra `<word>' en castellano? Responde solo `Sí' o `No' y sé sincero, por favor.}'' (``Does the word `<word>' exist in Spanish? Just answer `Yes' or `No' and be honest, please.'')
    \item \textbf{Prompt 3}: ``\textit{¿Es correcta la palabra `<word>' en castellano? Responde solo `Sí' o `No' y sé sincero, por favor.}'' (``Is the word `<word>' valid in Spanish? Answer only `Yes' or `No' and be honest, please.'')
\end{itemize}

These three prompts will enable us to assess if the models can be trusted when using ``Yes/No'' prompts. This is relevant because if these prompts were reliable, then an automatic evaluation of for example all the words in the dictionary could be easily run.  

\subsection{Test words}

The best way to evaluate the knowledge of the words comprehensively seems to be to use all the words in a dictionary. Among the Spanish dictionaries, there are two that stand out above the rest. 
The first is the Diccionario de la lengua española (DRAE)\footnote{\url{https://dle.rae.es/}}, by the Real Academia Española. This dictionary records 93,000 words from the 18th century onwards, and it is widely regarded as the main reference and authority in the language \cite{martinez1995vox} (pp. 232-233). 

The second dictionary to consider is the Diccionario del español actual (DEA)\footnote{\url{https://www.fbbva.es/diccionario/info/el-diccionario/}}, by Seco, Andrés and Ramos \cite{seco1999diccionario}. The DEA is a compilation of 84,000 words used in contemporary Spanish since 1950. It is based on actual documents and complements the definitions with examples of the words in use. This dictionary is considered exceptionally comprehensive and consistent and has been described as the most important lexicographical work in centuries \cite{alvarez2011diccionarios} (pp. 141-150). 

As we wish to assess the model’s knowledge of the Spanish lexicon that is used by humans today, and to evaluate the knowledge of the words exhaustively, the DEA seems more appropriate for our objectives: it excludes outdated lexicon and presents detailed information about the current use of words. 

Ideally, we would like to test all the words in the dictionary. However, that is only possible if we can trust the model responses, something that is not generally the case \cite{hallucinationsReview}. Therefore, we start with a test set of 100 words that are randomly taken from the Diccionario del español actual and use them to conduct a manual evaluation as described next\footnote{The list of the words is included in section \ref{Annexdata} and the dataset with the raw data from the 100 words  in \url{https://zenodo.org/doi/10.5281/zenodo.10797991} }.


\subsection{Procedure}

The models were loaded into our computing cluster and the five prompts were run sequentially for the 100 test words, setting the parameters of the models to obtain reproducible and deterministic results\footnote{This was done by disabling sampling or setting the temperature to zero depending on the model.}.


The first step of the analysis was to check manually the answers to prompts A (meaning) and B (use). This was done by an expert on languages with over 20 years of experience teaching Spanish in different Universities in the UK and Asia as well as a professional interpreter and writer. The answer produced by each model is compared with that in the dictionary and marked as valid if there is a match. For words with several meanings, we accept as positive a response that contains at least one of the meanings as long as it does not include additional incorrect meanings. The use of the word in the sentences with Prompt B is also checked manually. This result is assumed to be the ground truth on the knowledge that the model has of a word.

To assess if an automatic analysis is possible, in a second step the manual results are compared with those of prompts 1,2,3 to see if there is a correlation between the model answers to ``Yes/No'' prompts and the ground truth.  The idea is to check if having different responses on the prompts for the same word can be used as an indication that the responses are false positives and negatives. This would enable the evaluation of all the words in the dictionary, something that cannot be checked manually in a reasonable time. 

Finally, an alternative approach to automate the analysis is evaluated by using ChatGPT to check the model responses. This is done 1) by asking ChatGPT if the meaning given by a model corresponds to the word and 2) by asking ChatGPT if the meaning given by a model matches the one in the dictionary for that word. In more detail, these checking prompts used are:

\begin{itemize}
    \item \textbf{Check 1}: ``\textit{Es correcta la definición <definition> para la palabra <word>? Responde solo sí o no}'' (``Is the definition <definition> for the word <word> correct? Answer yes or no only.'')
    \item \textbf{Check 2}: ``\textit{¿Son equivalentes las siguientes definiciones?  <1. {definition}> <2. {definition}>.}'' (``Are the following definitions equivalent? <1. {definition}> <2. {definition}>'') with the first definition being the one in the dictionary and the second the one given by the model.
\end{itemize}

again, the responses to those checking prompts will be correlated to the ground truth to see if the automatic checks are reliable.

\section{Results} 
\label{sec:results}

In this section, the results for the manual testing are presented first to then analyze the feasibility of using the ``Yes/No'' prompts or the checking by ChatGPT to automate the process. 

\subsection{Manual evaluation}

We start by presenting the summary of the results for the manual analysis in Table \ref{tab:summary}. It can be seen that all models fail to produce a valid meaning for a significant fraction of the words and are unable to construct meaningful sentences using the words in context for most of the words. This clearly shows that there is room for improvement in the lexical knowledge of Spanish in open conversational LLMs. Analyzing the results in more detail, the following observations can be made:

\begin{enumerate}
    \item \textbf{Observation 1: Valid meanings are below 50\%.} Two-thirds of the models do not produce valid meanings for more than half of the words evaluated. Only one-third of the models can reach 50\% with the best achieving only 66\%.
    \item \textbf{Observation 2: Correct word usage is below 25\%.} Only one model reaches 25\% correct usage of the words and the majority of the models are below 10\%. 
    \item \textbf{Observation 3: Performance improves with model size.} Both meaning and usage increase with model size for Llama, Mistral, and Yi, suggesting that larger models can better handle the lexicon.
    \item \textbf{Observation 4: Adaptation to Spanish does not improve performance.} Models that have been designed to support multiple languages (such as Bloomz) or that have been adapted or fine-tuned for Spanish (Flor and Bertin) have lower scores in both meanings and usage than Llama or Mistral models of the same size. 
\end{enumerate}

\begin{table*}[h]
    \centering
    \begin{tabular}{|c|c|c|}
    \hline
    Model & Word meaning & Word use \\
    \hline
    Llama-2-7b-chat-hf & 42  & 3   \\ 
    Llama-2-13b-chat-hf & 46 &  20  \\
    Llama-2-70b-chat-hf & \textbf{50}  &  25 \\ 
    \hline
    Mistral-7b-Instruct & \textbf{51} &  6 \\ 
    Mixtral-8x7b-Instruct  & \textbf{66} & 17  \\ 
    \hline
    Gemma-7b-it & 20 & 4   \\ 
    \hline    
    SOLAR-10.7B-Instruct & \textbf{59} &  11 \\
    \hline    
    Yi-6B-Chat & 27 &  2  \\ 
    Yi-34B-Chat & 45 &  16  \\ 
     \hline    
    Bloomz-7b1  & 38 & 0  \\
    \hline    
    FLOR-6.3b-Instructed  & 39 & 0 \\ 
    \hline    
    Bertin-6b & 19 & 4   \\ 
    \hline    
    \end{tabular}
     \caption{Number of correct answers in the manual evaluation (in bold values above 50\%).}
    \label{tab:summary}
\end{table*}

Looking at the results for each model individually, the following observations can be made which show limitations of the models and a tendency towards English:

\begin{enumerate}
    \item Responses to prompt A (definition) and prompt B (language in use) are not necessarily in line; i.e., the meaning that the definition refers to and the text produced are frequently not consistent. For example, ‘tundidor’ is defined as a shy person in the answer to prompt A, but the word is used for describing a type of machine in prompt B (Bloomz-7b1); ‘sagum’ is long hair in the definition, but a piece of cloth in the text produced (Mistral-7b); etc. This behaviour is frequent across all models. 
    \item Grammatical mistakes are rare, but when these occur, they are related exclusively to agreements (masculine/feminine, singular/plural: ``la olor'' \textcolor{black}{[``la'' is femenine and ``olor'' is masculine]}, ``que no se pierdan nadie'' \textcolor{black}{[``pierdan'' is the verb form for plural, but for ``nadie'' the singular form should be used]}, ``los sílabas'' \textcolor{black}{[``los'' is masculine and ``sílabas'' is femenine]}, ``el yema'' \textcolor{black}{[``el'' is masculine and ``yema'' is femenine]}…).
    \item Structures are heavily influenced by English, which results in unusual constructions (``finalmente fue completado'' \textcolor{black}{[finally completed]}, ``socialmente responsable'' \textcolor{black}{[socially responsible]}, ``era frecuentemente invitado'' \textcolor{black}{[was frequently invited]}…). 
    \item Most models tend to include English in their responses, producing answers fully or partially written in that language: ``no se puede \textcolor{black}{\textit{missed}}'' \textcolor{black}{(it can't be missed)}, ``prolongado \textcolor{black}{\textit{enough}} to be measurable'' \textcolor{black}{(extended enough to be measurable)}, ``una cola larga y \textcolor{black}{\textit{bushy}}'' \textcolor{black}{(a long and bushy tail)}, etc. Yi-34B also responds including Russian and Chinese words.
    \item The frequency of non-existent hybrid words mixing English and Spanish is high: ``Ella es meana'' (for ``she is mean''), ``hoefados'' (meaning ``with a hoef''), ``si meantes'' (``if you mean'')…  Most of these words are incomprehensible: ``oratoo'', ``cérama'', ``metingse'', ``lasufación'', ``conmanes''.
    \item All models are unable to identify the use of foreign words in Spanish. When asked about those terms, they answer directly in the language the word comes from, and refer to the meaning in that language alone (``speed'' is only ``fast pace'', not a drug; ``apparat'' is defined mechanically, not politically, etc.).
    \item Definitions are sometimes mere descriptions of a word’s features (for example, Bertin-6b just provides grammatical category and even number of letters). Many definitions refer to the (often incorrect) translation of the word, instead of describing its meaning (Yi-6b, for example, responds that ``to acelguilla'' [a plant] ``puede significar to adjust o to fit'' \textcolor{black}{[may mean to adjust or to fit]}, ``ardilla'' [squirrel] ``significa rabbit en inglés'' \textcolor{black}{[means rabbit in English]}; etc.
    \item Some models (Llama-2-13b, Llama-2-7b) provide unsolicited etymological information which is mostly inaccurate: `` `verrax' means urine in Latin'', `` `colosseum' comes from the Latin `column' ''; `` `cosmo' is a prefix referring to a unit of measurement''; etc.).
    \item Common words (``minute'', ``tie'', ``cent'', ``thirty'') are often defined incorrectly. 
\end{enumerate}

\begin{figure}[h]
  \centering
  \includegraphics[scale=0.55]{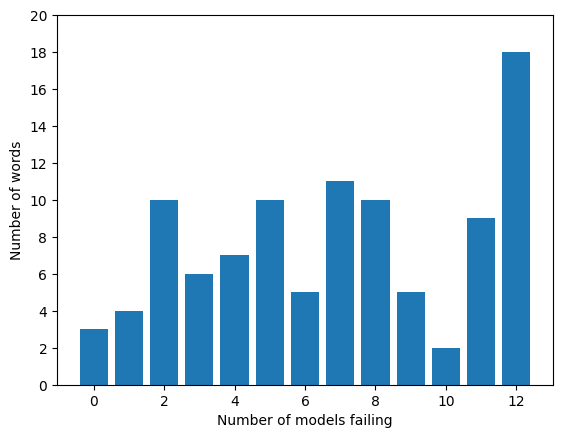}
  \caption{Number of words failing on a number of models.}
  \label{fig:ModelFail}
\end{figure}

\begin{figure}[h]
  \centering
  \includegraphics[scale=0.55]{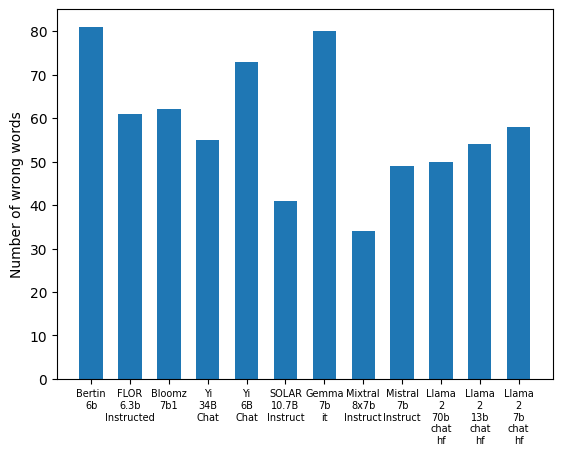}
  \caption{Number of words failing on each model.}
  \label{fig:WordsModel}
\end{figure}

It is also of interest to study if the errors made by the models are concentrated on the same words or if the errors are more dependent on the model with some models failing in some words and others in different words. Figure \ref{fig:ModelFail} shows the number of words that fail on a given number of models. It can be seen that the maximum occurs for failures on all (twelve) models but there are also words in each of the other categories. Therefore, there is a correlation between the failures of the models but not in all cases. The number of words failing per model is shown in Figure \ref{fig:WordsModel} showing also significant variability among models. The results for each word and model are included in the annex in section \ref{Annexdata}.

\begin{figure}[h]
  \centering
  \includegraphics[scale=0.55]{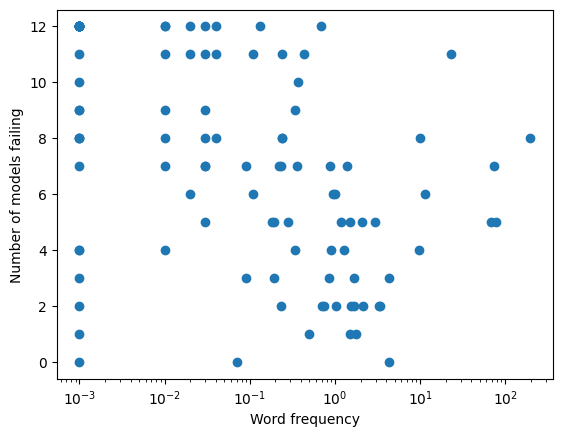}
  \caption{Number of models failing per word versus word frequency \scriptsize (to represent words that do not appear in CREA, we plot them with a frequency of $10^{-3}$  as the frequency is represented in a logarithmic scale.).}  
  \label{fig:Freq}
\end{figure}

To further analyze the words that are not correctly identified by the models, we checked the frequencies of the 100 test words in the Corpus de Referencia del Español Actual (CREA)\footnote{Real Academia Española: Banco de datos (CREA) Corpus de referencia del español actual \url{https://corpus.rae.es/lfrecuencias.html} consultado en febrero de 2024.}. Figure \ref{fig:Freq} shows the words as points in a plot with the word frequency on the x-axis and the number of failures in the 12 models on the y-axis. Most failures would be expected for words with low frequencies that are less used and thus will most likely appear fewer times in the training datasets. Therefore, a strong correlation between frequency and failures would be expected with the points concentrating towards zero (no failures) as frequency increases. The 100-word frequency and error distributions do not fit the Pearson assumptions, so the Spearman correlation was computed. The results suggest a statistically significant moderate negative monotonic relationship between the variables ($\rho$ = -.47, p-value < .0001).   However, there is a significant number of models that do not produce a valid meaning for the most common words in the test set. For example, the word with the largest frequency, ``minuto'' fails in the meaning for eight of the twelve models. This suggests that the presence of the word in the training dataset does not guarantee that the model will be able to produce a valid meaning for it and further shows the limitation of the models in Spanish. 

\subsection{Automation feasiblity}

After analyzing the manual results we consider whether the ``Yes/No'' prompts or the checks by ChatGPT can be used to automate the evaluation of the lexical knowledge so that all words in the dictionary can be evaluated. To do so we compare the results of prompts 1 to 3 and checks 1 and 2 with the ground truth using ChatGPT4 and show the accuracy for each of them. The results are summarized in table \ref{tab:automation}. It can be seen that the "Yes/No" prompts have poor accuracy for most models. In fact, the models that have the highest accuracy, for example Gemma with 80\% on prompts 1 to 3, answer ``No'' for all words which provides no useful information on the knowledge of words. Therefore, the ``Yes/No'' prompts cannot be used to conduct evaluation at scale as the results would not be reliable. The accuracy for the ChatGPT-based checks is better, specially for check 2 that reaches in some cases 90\%. However, even for these checks accuracy is below 70\% for some models. Therefore, overall the values are not still high enough to be used for testing at scale.

\begin{table*}[h]
   \scriptsize
    \centering
    \begin{tabular}{|c|c|c|c|c|c|}
    \hline
    Model & P1 (\%) & P2 (\%) & P3 (\%) & C1 (\%) & C2 (\%) \\
    \hline
    Llama-2-7b-chat-hf & 60 & 64 & 67  & 65  & 81    \\
    Llama-2-13b-chat-hf & 56 & 53 & 54 & 73 &  90   \\ 
    Llama-2-70b-chat-hf & 52 & 62 & 54 & 72 &  86  \\ 
    \hline
    Mistral-7b-Instruct & 53 & 66 & 74 & 82 & 86    \\ 
    Mixtral-8x7b-Instruct & 38 & 39 & 38 & 70 &  86   \\ 
    \hline
    Gemma-7b-it & 80 & 80 & 80 & 56 & 57   \\ 
    \hline    
    SOLAR-10.7B-Instruct & 56 & 48 & 48  & 72  & 87    \\ 
    \hline    
    Yi-6B-Chat & 48 & 73 & 75 & 74 & 76    \\ 
    Yi-34B-Chat & 39 & 51 & 59 & 62  &  77   \\ 
     \hline    
    Bloomz-7b1 & 38 & 38 & 38 & 77 & 75   \\     
    \hline    
    FLOR-6.3b-Instructed & 66 & 62  & 61 & 67 & 67   \\ 
    \hline    
    Bertin-6b & 81 & 77 & 19 & 73 & 67   \\ 
    \hline    
    \end{tabular}
    \caption{Accuracy of the "Yes/No'' checks (P1,P2,P3) and ChatGPT4 checks (C1,C2).}
    \label{tab:automation}
\end{table*}

\section{Discussion}
\label{sec:discussion}

The evaluation results show that open conversational LLMs have limited knowledge of the Spanish lexicon. This could be expected as most models have been trained with datasets that are dominated by English or Chinese. However, similar results are observed for models that have been designed to target many languages or fine-tuned for Spanish. This suggests that current efforts to support Spanish in LLMs are not effective in terms of lexical knowledge. \textcolor{black}{In fact, it would be of interest to do the same evaluation in English and compare the results as we are taking for granted that performance in English will be better.}

The evaluation also shows that automating the evaluation of lexical knowledge is not straightforward. This is a serious limitation as manual checking is only practical for a reduced set of words. \textcolor{black}{Further work is needed to improve automated testing, for example analyzing the mistakes made by ChatGPT in making the judgements may provide insights on how to improve the checking.}

In fact, the results presented in this paper for 100 words provide a first estimate of lexical knowledge but more words, \textcolor{black}{ideally thousands,} should be evaluated to have better estimates. Another limitation of our work is that results may depend on the prompts used, several prompts were tested and the differences observed were not large. However, the use of additional prompts to check the consistency of the results is also of interest. \textcolor{black}{The use of more advanced prompting techniques such as chain of thoughts would also be of interest, especially when the models are asked about their knowledge of words as LLMs have limited meta-congnition capabilities.} Again, having to conduct a manual analysis of 100 words on 12 models makes any additional testing cumbersome. \textcolor{black}{Another limitation of our study is that only three multilingual models have been tested as there are few such models. As newer multilingual models appear, it would be interesting to see if they improve their lexical knowledge of Spanish.}

Despite the limitations of our study, the limited lexical knowledge of Spanish in the open conversational LLM ecosystem is evident, \textcolor{black}{even for the multilingual models evaluated.} Most likely, the same applies to other languages that have even less presence on the training datasets. Therefore, an effort should be made by the open-source community to develop conversational LLMs with better lexical knowledge of Spanish. This is important given the large number of native speakers of Spanish and the expected impact of conversational LLMs in the future of languages.

\section{Conclusion}
\label{sec:conclusions}

In this paper, we have analyzed the lexical knowledge that open conversational LLMs have of Spanish. The results show that most LLMs fail to recognize more than half of the words tested and are unable to use most of the words in context when asked to do so with a simple prompt. This evaluation is done on a small sample of one hundred words randomly taken from a reference Spanish dictionary and thus is only an initial estimate of the lexical knowledge. However, even when taking that into account, the results show the limitation of open conversational LLMs in using the Spanish lexicon. This applies even to models that have been adapted or fine-tuned for Spanish. 

The automation of the lexical evaluation at scale has also been studied by using prompts with binary answers that can be processed automatically or by using a commercial conversational LLM as the evaluator judging if the responses are valid or comparing the meanings of the words produced by the LLM tested with those in the dictionary. The results show that automating the testing incurs a significant loss of accuracy as the LLMs produce false results for binary prompts frequently and the checking by another conversational LLM also has a relevant number of errors. Developing techniques to automate the testing of words is interesting as it would enable the evaluation of for example all the words in the dictionary.

The initial study presented in this paper can be extended by considering additional LLMs, a larger set of test words, and different prompts for the evaluation. Another interesting topic is the development of automated tests for lexical knowledge that overcome the challenges described in this paper. More generally and more importantly, our results suggest that the open-source LLM community should make an effort to better support languages other than English.

\section{Acknowledgements}


This work was supported by the FUN4DATE (PID2022-136684OB-C21/C22) project funded by the Spanish Agencia Estatal de Investigacion (AEI) 10.13039/501100011033, by the Chips Act Joint Undertaking project SMARTY (Grant no. 101140087) and by the OpenAI Researcher Access Program. The evaluation was also done in part with equipment that was donated by NVIDIA to support our research.

\bibliographystyle{fullname}


\bibliography{SpaDic}

\clearpage

\onecolumn

\appendix
\section{Annex: Test words and results}
\label{Annexdata}
\begin{figure*}[h]
  \centering
  \includegraphics[scale=0.65]{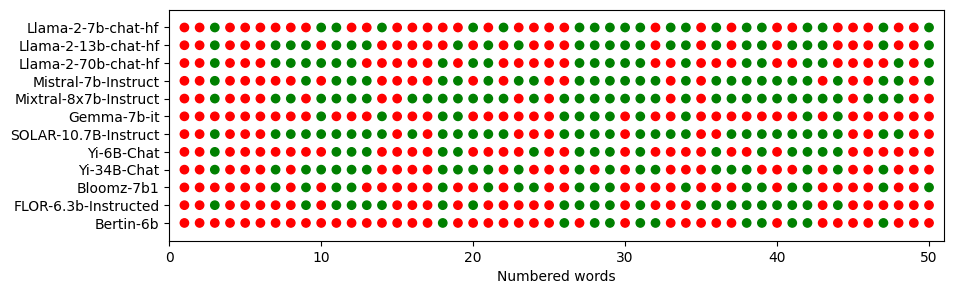}
  \includegraphics[scale=0.65]{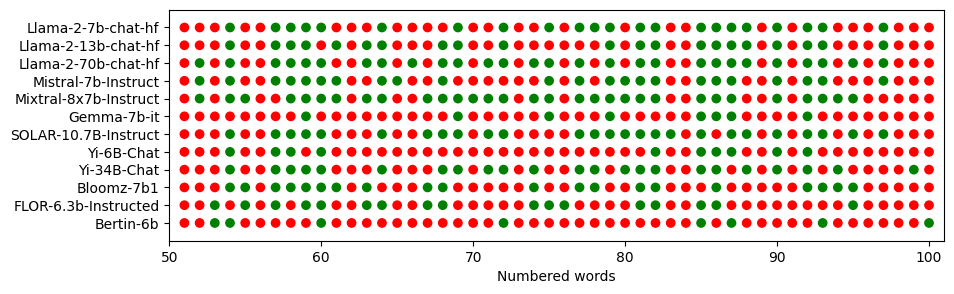}
  \caption{Models failing per word (red for failure, green for correct meaning).}
  \label{fig:WordsDot}
\end{figure*}

\begin{figure*}[h]
  \centering
  \includegraphics[scale=0.65]{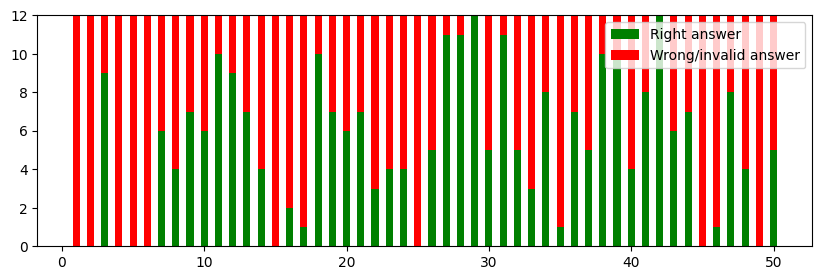}
  \includegraphics[scale=0.65]{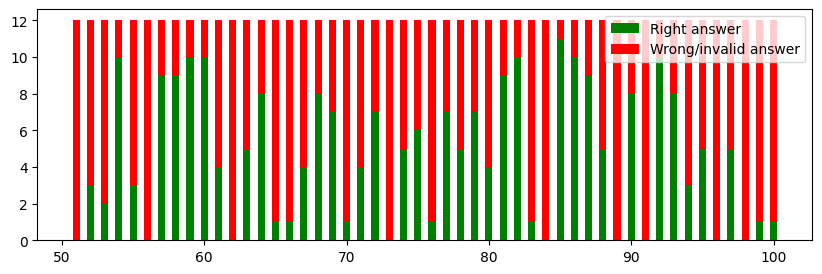}
  \caption{\textcolor{black}{Number of models failing per word (red for failure, green for correct meaning).}}
  \label{fig:WordsBar}
\end{figure*}

The words used in the evaluation are listed in Table \ref{tabla:palabras7} and the results for each of them for each model are shown in Figure \ref{fig:WordsDot} \textcolor{black}{with the number of models failing in Figure \ref{fig:WordsBar}}. Additional details and the text generated by the models are available in the public repository.  


\begin{table*}[h]
\scriptsize
\centering
\begin{tabular}{|*{7}{l|}}
\hline
\textcolor{red}{acelguilla} & \textcolor{red}{agüista} & \textcolor{ForestGreen}{antidiarreico} & \textcolor{red}{apealar} & \textcolor{red}{apparat} & \textcolor{red}{arante} & \textcolor{orange}{ardilla} \\
\hline
\textcolor{orange}{bátavo} & \textcolor{orange}{bicicross} & \textcolor{orange}{bifocal} & \textcolor{ForestGreen}{cantautor} & \textcolor{ForestGreen}{cantinero} & \textcolor{orange}{cartulina} & \textcolor{orange}{centavo} \\
\hline
\textcolor{red}{cerebrotónico} & \textcolor{red}{chalaza} & \textcolor{red}{chigüire} & \textcolor{ForestGreen}{coliseo} & \textcolor{orange}{conspirar} & \textcolor{orange}{corbata} & \textcolor{orange}{corralero} \\
\hline
\textcolor{red}{cosmotrón} & \textcolor{orange}{crístico} & \textcolor{orange}{cuentapartícipe} & \textcolor{red}{dabuti} & \textcolor{orange}{dactiloscopia} & \textcolor{ForestGreen}{deformabilidad} & \textcolor{ForestGreen}{desinfección} \\
\hline
\textcolor{ForestGreen}{desinsectación} & \textcolor{orange}{desmitificador} & \textcolor{ForestGreen}{diligentemente} & \textcolor{orange}{emparejador} & \textcolor{red}{empurpurado} & \textcolor{ForestGreen}{epifito} & \textcolor{red}{escorar} \\
\hline
\textcolor{orange}{estadista} & \textcolor{orange}{esteatosis} & \textcolor{ForestGreen}{estuco} & \textcolor{ForestGreen}{exequias} & \textcolor{orange}{faisánido} & \textcolor{ForestGreen}{fétido} & \textcolor{ForestGreen}{floración} \\
\hline
\textcolor{orange}{fotogramétrico} & \textcolor{orange}{funcionario} & \textcolor{red}{gabato} & \textcolor{red}{garcilla} & \textcolor{orange}{giroscopio} & \textcolor{orange}{helenizante} & \textcolor{red}{hipogino} \\
\hline
\textcolor{orange}{incrustar} & \textcolor{red}{intercadencia} & \textcolor{red}{jaín} & \textcolor{red}{lipotimia} & \textcolor{ForestGreen}{magnolia} & \textcolor{red}{manes} & \textcolor{red}{meano} \\
\hline
\textcolor{ForestGreen}{mediar} & \textcolor{ForestGreen}{mensualizar} & \textcolor{ForestGreen}{mesmerización} & \textcolor{ForestGreen}{mestizo} & \textcolor{orange}{minuto} & \textcolor{red}{mochalero} & \textcolor{orange}{modisto} \\
\hline
\textcolor{ForestGreen}{monásticamente} & \textcolor{red}{morra} & \textcolor{red}{nefólogo} & \textcolor{orange}{novatada} & \textcolor{ForestGreen}{ovni} & \textcolor{orange}{pagar} & \textcolor{red}{paleteo} \\
\hline
\textcolor{orange}{palmítico} & \textcolor{orange}{paralogismo} & \textcolor{red}{pasarratos} & \textcolor{orange}{perrillo} & \textcolor{orange}{pezuña} & \textcolor{red}{pinabete} & \textcolor{orange}{pitahaya} \\
\hline
\textcolor{orange}{postrar} & \textcolor{orange}{prédica} & \textcolor{orange}{prolongador} & \textcolor{ForestGreen}{provinciano} & \textcolor{ForestGreen}{puzzle} & \textcolor{red}{quepis} & \textcolor{red}{raor} \\
\hline
\textcolor{ForestGreen}{reciclado} & \textcolor{ForestGreen}{rememorar} & \textcolor{ForestGreen}{ridiculización} & \textcolor{orange}{sagum} & \textcolor{red}{salbanda} & \textcolor{ForestGreen}{socialmente} & \textcolor{red}{speed} \\
\hline
\textcolor{ForestGreen}{standarizar} & \textcolor{ForestGreen}{sublimado} & \textcolor{red}{superbomba} & \textcolor{orange}{talgo} & \textcolor{red}{tornajo} & \textcolor{orange}{treinta} & \textcolor{red}{tundidor} \\
\hline
\textcolor{red}{vega} & \textcolor{red}{verraquear} & & & & & \\
\hline
\end{tabular}
\caption{Subset of words used in the evaluation. \textcolor{black}{Those failing in nine or more models are in red, those failing in four or fewer models in ForestGreen and the rest in orange.}}
\label{tabla:palabras7}
\end{table*}

\end{document}